\documentclass[lettersize,journal]{IEEEtran}
\usepackage{amsmath,amsfonts}
\usepackage{array}
\usepackage{textcomp}
\usepackage{stfloats}
\usepackage[breaklinks,colorlinks]{hyperref}
\usepackage{url}
\usepackage{verbatim}
\usepackage{graphicx}
\usepackage{cite}
\usepackage{makecell}
\usepackage{tabulary}
\usepackage{cases}
\usepackage{multirow}
\usepackage[dvipsnames]{xcolor}
\usepackage{booktabs}
\usepackage{colortbl}

\usepackage{algorithm}
\usepackage{algpseudocode}
\usepackage{newtxmath}

\usepackage{float}
\usepackage{booktabs}
\usepackage{tabulary}
\usepackage{cuted}
\usepackage{caption}
\usepackage{subfig}

\newcommand{\tablestyle}[2]{\setlength{\tabcolsep}{#1}\renewcommand{\arraystretch}{#2}\centering\footnotesize}
\makeatletter
\newcommand{\ie}{\emph{i.e.}\@ifnextchar.{\!\@gobble}{}}
\newcommand{\eg}{\emph{e.g.}\@ifnextchar.{\!\@gobble}{}}
\newcommand{\etc}{etc\@ifnextchar.{}{.\@}}
\makeatother

\begin{document}

% \title{UHR-DETR: Efficient End-to-End Small Object Detection in UHR Remote Sensing Imagery}
\title{UHR-DETR: Efficient End-to-End Small Object Detection for Ultra-High-Resolution Remote Sensing Imagery}

\author{Jingfang~Li,~
\and Haoran~Zhu,~
\and Wen~Yang,~\IEEEmembership{Senior~Member,~IEEE,}~
\and Jinrui~Zhang,~
\and Fang~Xu, 
\and Haijian~Zhang,~\IEEEmembership{Senior~Member,~IEEE,}
\and Gui-Song Xia,~\IEEEmembership{Senior Member,~IEEE}
% stops a space
\thanks{J. Li, H. Zhu, W. Yang, J. Zhang, and H. Zhang are with the School of Electronic Information, Wuhan University, Wuhan, 430072 China. \emph{E-mail: \{lijingfang, zhuhaoran, yangwen, jrzhang, haijian.zhang \}@whu.edu.cn}}
% \thanks{C. Xu is with the Environmental Computational Science and Earth Observation Laboratory, EPFL, Sion, Switzerland. \emph{E-mail: chang.xu@epfl.ch}}
% \thanks{F. Xu and G-S. Xia are with the School of Artificial Intelligence, Wuhan University, Wuhan, 430072, China. \emph{E-mail: xufang, guisong.xia@whu.edu.cn}}
\thanks{F. Xu and G-S. Xia are with the School of Artificial Intelligence, Wuhan University, Wuhan, 430072, China. \emph{E-mail: \{xufang, guisong.xia\}@whu.edu.cn}}
\thanks{Jingfang Li and Haoran Zhu contributed equally to this work.}
\thanks{Corresponding author: Wen Yang.}
}

% The paper headers
\markboth{Journal of \LaTeX\ Class Files,~Vol.~14, No.~8, August~2021}%
{Shell \MakeLowercase{\textit{et al.}}: A Sample Article Using IEEEtran.cls for IEEE Journals}

% \IEEEpubid{0000--0000/00\$00.00~\copyright~2021 IEEE}
% Remember, if you use this you must call \IEEEpubidadjcol in the second
% column for its text to clear the IEEEpubid mark.

\maketitle

\begin{strip}
    \centering
    \vspace{-80pt}
    \includegraphics[width=\linewidth]{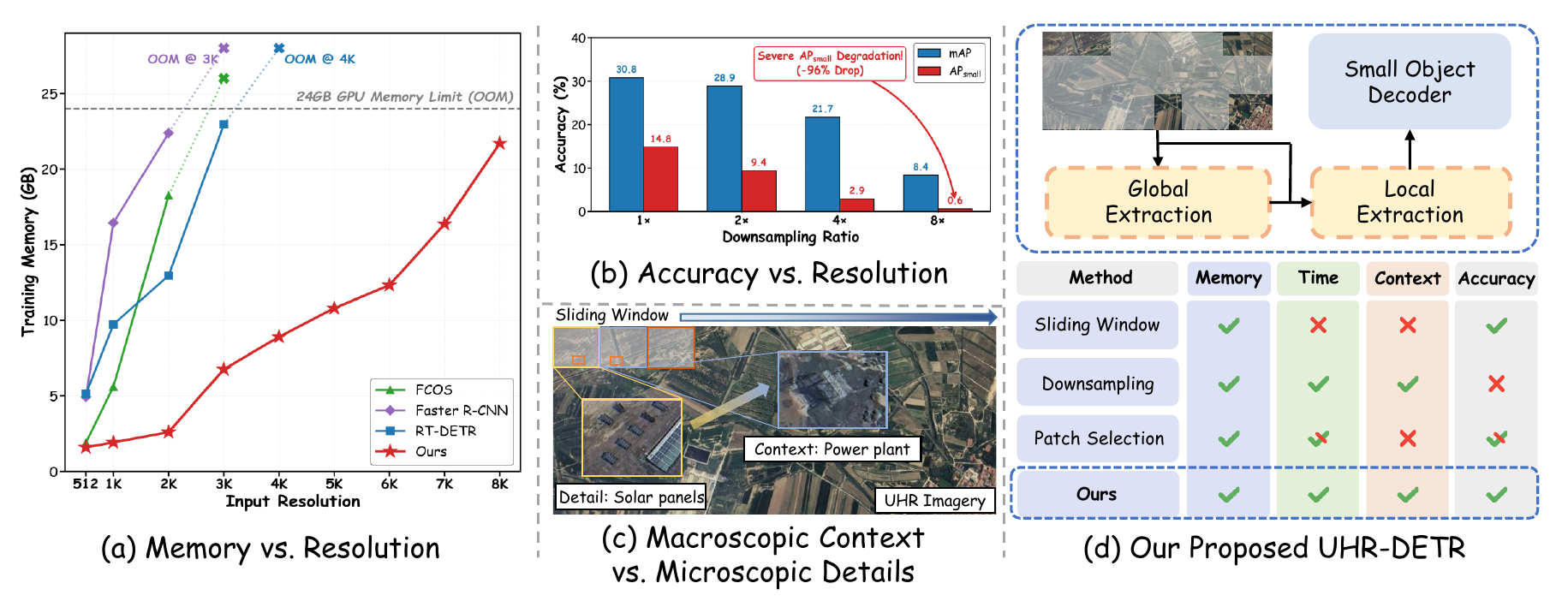}
    % \captionof{figure}{Overview of our study on generalized small object detection. Leveraging the proposed TinySet-9M dataset and benchmark, we systematically investigate the performance of existing label-efficient paradigms in the small-object regime and introduce a new detection paradigm, Point-prompt Small Object Detection (P$^2$SOD). The middle panel illustrates the domain composition of TinySet-9M, while the right panel compares the performance of representative label-efficient paradigms on small objects and our proposed detection paradigm.}
    \captionof{figure}{\textbf{The perception paradox in UHR imagery and our solution.} \textbf{(a)} Standard object detectors suffer from severe Out-of-Memory (OOM) bottlenecks on 24GB GPUs, whereas our method efficiently scales up to ultra-high resolutions. \textbf{(b)} Naive downsampling leads to a catastrophic drop in small object accuracy. \textbf{(c)} Sliding window strategies inevitably fracture crucial macroscopic context. \textbf{(d)} To break these dilemmas, UHR-DETR introduces a global-local decoupled paradigm, achieving an optimal balance among memory efficiency, inference speed, context preservation, and detection accuracy.}
    \label{fig:first_figure}
    \vspace{0pt}
\end{strip}

\begin{abstract}
Ultra-High-Resolution (UHR) imagery has become essential for modern remote sensing, offering unprecedented spatial coverage.
However, detecting small objects in such vast scenes presents a critical dilemma: retaining the original resolution for small objects causes prohibitive memory bottlenecks. Conversely, conventional compromises like image downsampling or patch cropping either erase small objects or destroy context.
To break this dilemma, we propose UHR-DETR, an efficient end-to-end transformer-based detector designed for UHR imagery. First, we introduce a Coverage-Maximizing Sparse Encoder that dynamically allocates finite computational resources to informative high-resolution regions, ensuring maximum object coverage with minimal spatial redundancy. Second, we design a Global-Local Decoupled Decoder. By integrating macroscopic scene awareness with microscopic object details, this module resolves semantic ambiguities and prevents scene fragmentation. Extensive experiments on the UHR imagery datasets (\textit{e.g.}, STAR and SODA-A) demonstrate the superiority of UHR-DETR under strict hardware constraints (\textit{e.g.}, a single 24GB RTX 3090). It achieves a 2.8\% mAP improvement while delivering a 10$\times$ inference speedup compared to standard sliding-window baselines on the STAR dataset. Our codes and models will be available at \href{https://github.com/Li-JingFang/UHR-DETR}{GitHub}.

\end{abstract}


\section{Introduction}
\label{introduction}

% revised by Zhu
\IEEEPARstart{D}{etecting} small objects in remote sensing imagery has achieved remarkable progress~\cite{tgrs_ffcayolo, tgrs_bafnet}, serving as a cornerstone for macroscopic applications such as urban planning and maritime surveillance. However, driven by the rapid development of modern Earth observation systems and the growing demand for continuous wide-area monitoring, Ultra-High-Resolution (UHR) imagery is rapidly establishing itself as the new standard for macroscopic perception~\cite{star_2025_tpami}. For instance, given a standard Ground Sample Distance (GSD) of 0.5~m/pixel, DOTA benchmark~\cite{DOTA2.0_2021_tpami} images (\textit{e.g.}, $1024 \times 1024$) cover a highly restricted physical area of merely 0.26~km$^2$. In stark contrast, a UHR image easily spans $8192 \times 8192$ pixels~\cite{geollava8k_2025_nips}, encompassing an extensive area of nearly 17~km$^2$. This leap in spatial dimensions imposes unprecedented demands on object detectors. 
The challenge is further exacerbated by the fact that perceiving small objects requires preserving fine-grained, high-resolution features to handle their densely packed distributions and the extreme scale disparity against the vast UHR canvas. Consequently, maintaining these high-resolution representations across such expansive spatial extents inevitably triggers an astronomical surge in memory footprint and severe bottlenecks in inference speed.

%On the one hand, perceiving small objects inherently requires preserving fine-grained, high-resolution features; when applied across such vast spatial extents, this inevitably triggers an astronomical surge in computational resource consumption and memory footprint. On the other hand, the exorbitant input resolution poses severe bottlenecks for inference speed, while the extreme scale variation and the presence of densely packed small objects make accurate and efficient perception difficult.

To alleviate the computational burden of UHR imagery, existing UHR object detection strategies generally fall into two categories. The first approach utilizes efficient non-uniform downsampling strategies to preserve a larger pixel ratio for foreground regions~\cite{fovea_2021_iccv, lzu_2023_cvpr, zoomdet_2026_isprs}. Alternatively, the second approach introduces selective filtering mechanisms into the standard sliding-window framework to skip empty backgrounds and improve detection efficiency~\cite{clusdet_2019_iccv, spdet_2023_icme}. Yet, despite the progress made over naive downsampling and exhaustive cropping, detecting small objects in massive scenes still faces profound challenges. The first challenge is the \textbf{\textit{severe conflict between resolution and efficiency}}: precise detection of small objects (under $32 \times 32$ pixels~\cite{coco_2014_eccv}) inherently requires high-resolution features, making it incredibly difficult to efficiently extract such features from UHR images (over $8192 \times 8192$ pixels) without exhausting computational resources.
As shown in Fig.~\ref{fig:first_figure}(b), increasing the downsampling ratio leads to a catastrophic 96\% relative drop in small object accuracy.
Secondly, the robust perception of small objects heavily relies on sufficient contextual support to alleviate semantic ambiguity. Since sliding window strategies inevitably \textbf{\textit{fracture the global image topology}}, effectively reconstructing and bridging this severed macroscopic context to aid microscopic detection remains a critical unsolved obstacle. 
As shown in Fig.~\ref{fig:first_figure}(c), rigid sliding windows isolate microscopic details (\textit{e.g.}, solar panels) from their corresponding macroscopic context (\textit{e.g.}, the power plant), depriving the detector of crucial semantic cues.

Motivated by these challenges, and building upon the advanced RT-DETR~\cite{rt_detr_2024_cvpr} architecture, we propose UHR-DETR, the first efficient, end-to-end transformer-based detector designed for UHR imagery. As illustrated in Fig.~\ref{fig:first_figure}(a), UHR-DETR bypasses the prohibitive memory and latency bottlenecks that paralyze standard models processing massive inputs. Moreover, it resolves the two aforementioned dilemmas through two core innovations. 
% First, to tackle the severe conflict between resolution and efficiency, we introduce a Coverage-Maximizing Sparse Encoder. Instead of exhaustively processing the entire image, this module leverages a lightweight backbone to extract global features and formulates spatial feature routing as a geometric set cover problem. Optimized via a novel Local Peak Margin Loss and executed through an Iterative Soft-Subtraction Greedy Algorithm, it dynamically allocates finite computational resources strictly to informative high-resolution regions.
First, to tackle the severe conflict between resolution and efficiency, we draw inspiration from the 'glimpse-and-fixate' mechanism of the human visual system. Just as humans naturally leverage a macroscopic glimpse to direct high-resolution foveal attention exclusively toward critical localized regions, we introduce a Coverage-Maximizing Sparse Encoder to mimic this biological paradigm. Instead of exhaustively processing the entire image, this module leverages a lightweight backbone to extract global features and formulates spatial feature routing as a geometric set cover problem. Optimized via a novel Local Peak Margin Loss and executed through an Iterative Soft-Subtraction Greedy Algorithm, it dynamically allocates finite computational resources strictly to informative high-resolution regions.
By doing so, the encoder effectively extracts the crucial regional features containing objects of interest, significantly reducing spatial redundancy by bypassing vast uninformative backgrounds.
Second, to reconstruct the severed holistic context, we design a Global-Local Decoupled Decoder. To circumvent the compromises of conventional patch inference and aggressive downsampling, our decoder introduces a progressive global-to-local dual cross-attention paradigm. Object queries initially interact with global features to establish macroscopic scene awareness, and are subsequently routed to the sparse high-resolution patches to perceive microscopic object details. This decoupled design explicitly integrates global topology with local refinement, facilitating robust local feature discrimination through global contextual guidance.
Extensive experiments on two massive benchmarks, STAR~\cite{star_2025_tpami} ($8192 \times 8192$) and SODA-A~\cite{soda_2023_tpami} ($9600 \times 9600$), demonstrate the superiority of UHR-DETR in handling densely packed small objects. Notably, our method achieves a 2.8 mAP improvement and over a 10$\times$ inference speedup compared to the exhaustive sliding-window RT-DETR. Moreover, it consistently surpasses existing state-of-the-art UHR detectors by a significant margin of 4.2 mAP, all while maintaining faster processing speeds.

Our main contributions are summarized as follows:
\begin{itemize}
    \item We propose UHR-DETR, the first efficient, end-to-end transformer-based detector specifically designed for small object detection in UHR remote sensing imagery. It addresses severe memory and computational bottlenecks, providing a holistic alternative to the traditional fragmented crop-and-detect paradigm.
    \item We introduce a detection architecture featuring a Coverage-Maximizing Sparse Encoder and a Global-Local Decoupled Decoder. This design dynamically routes finite computation to high-value regions via a geometric set cover formulation, and seamlessly fuses macroscopic scene context with microscopic details.
    \item Extensive experiments on the STAR and SODA-A datasets validate the effectiveness of UHR-DETR. Compared to standard baselines, our method maintains comparable detection accuracy while achieving over a $10\times$ reduction in inference latency, successfully processing massive UHR inputs and overcoming the severe memory limitations inherent in standard architectures.
\end{itemize}

The remainder of this paper is organized as follows. First, we review related work on transformer-based small object detection and ultra-high-resolution object detection in Section~\ref{related_work}. Next, we detail the proposed UHR-DETR framework, focusing on the Coverage-Maximizing Sparse Encoder and the Global-Local Decoupled Decoder, in Section~\ref{methods}. Subsequently, we present extensive experiments and comprehensive ablation studies conducted on multiple public datasets in Section~\ref{experiments}. Additionally, we discuss the underlying design principles and inherent limitations of our approach in Section~\ref{discussion}. Finally, we conclude the paper in Section~\ref{conclusion}.

\section{Related Work}
\label{related_work}

% \subsection{Small Object Detection}

% Detecting small objects remains a formidable challenge in remote sensing community. Their scarce visual features and high susceptibility to background noise make them extremely difficult to localize and classify accurately.

% To systematically benchmark algorithms in this domain, evaluation datasets have progressively evolved. They transitioned from general-purpose benchmarks (\textit{e.g.}, MS-COCO~\cite{coco_2014_eccv}, DIOR~\cite{dior_2020_isprs}) to dedicated datasets explicitly curated for small objects, including AI-TODv2~\cite{aitodv2_2022_isprs} and SODA~\cite{soda_2023_tpami}.

% From an algorithmic perspective, existing advancements primarily focus on feature enhancement, label assignment, and query optimization. Methods enrich semantics via feature pyramids \cite{fpn_2017_cvpr} or super-resolution \cite{mtgan_2018_eccv}, and improve label assignment using robust metrics like Wasserstein distance \cite{aitodv2_2022_isprs} to mitigate IoU sensitivity. Additionally, recent DETR variants dynamically adjust the number of object queries to better capture small instances \cite{dqdetr_2024_eccv, d3q_2025_jstar}.

% Despite these significant advancements, both current datasets and algorithms are predominantly tailored for standard-sized images or isolated local crops. While highly effective within limited local regions, this paradigm leaves the continuous wide-area context inherent in massive ultra-high-resolution scenes largely unexplored.

\subsection{Transformer-based Small Object Detection}
Detecting small objects has been a central focus in the remote sensing community. Owing to their ability to model long-range contextual dependencies, transformer-based architectures have become the dominant paradigm for small object detection. Existing Transformer-based detection methods can be broadly categorized into two directions: feature representation enhancement and learning strategies optimization.

Feature representation enhancement aims to leverage transformers' attention mechanisms to assign more discriminative features to small objects. Deformable DETR~\cite{deformable_detr_2021_iclr} first introduced Deformable Attention, which combines the sparse sampling capability of DCN with the global relational modeling of Transformers, significantly enhancing the representation of small objects over DETR. DART~\cite{DART-detr} further improves spatial perception for small objects by dynamically generating content-aware regions via an adaptive-region attention mechanism. Cross-DINO~\cite{cross-dino} incorporates a deep MLP to aggregate initial features with both short- and long-range information, strengthening representations for small objects. D$^3$R-DETR~\cite{d3r-detr} exploits frequency-domain features to better capture contextual cues for small objects.
Learning strategy optimization focuses on improving transformer queries and hungarian matching to enable faster querying and better convergence for small objects. Co-DETR~\cite{co-detr} introduces an auxiliary decoder that leverages dense label assignment (\textit{e.g.}, FCOS~\cite{fcos_2019_iccv}, ATSS~\cite{atss}) to enhance convergence. DQ-DETR~\cite{dqdetr_2024_eccv}, D3Q-DETR~\cite{d3q_2025_jstar}, Dome-DETR~\cite{dome-detr}, and DEAL~\cite{deal} generate query numbers from density maps (via classification or regression) and dynamically allocate queries during training, significantly improving performance on dense small object detection.

Although transformer-based methods for small object detection have achieved notable progress, their reliance on global contextual modeling leads to substantial memory consumption as image resolution increases. To address this limitation, we propose a Coverage-Maximizing Sparse Encoder that leverages sparse modeling to improve efficiency without sacrificing performance.

% However, a critical limitation remains: both current datasets and algorithms primarily focus on standard-sized images or isolated local crops. By restricting perception to limited local regions, they fundamentally neglect the continuous wide-area context inherent in massive ultra-high-resolution scenes.

\subsection{Ultra-High-Resolution Object Detection}

Driven by rapid advancements in sensor technologies, ultra-high-resolution imagery presents unique challenges for small object detection due to the extreme scale variation between vast backgrounds and densely packed small objects. 

To systematically benchmark algorithms in this domain, evaluation datasets have progressively expanded in spatial coverage. Early efforts primarily focused on standard high-resolution drone imagery, represented by datasets like VisDrone~\cite{visdrone_2021_tpami} and UAVDT~\cite{uavdt_2018_eccv}. However, to address wide-area perception, the community has transitioned to massive benchmarks, such as PANDA~\cite{panda_2020_cvpr} for gigapixel crowds and STAR~\cite{star_2025_tpami} for holistic satellite perception.
From an algorithmic perspective, various strategies have been proposed to alleviate the computational burden on these massive images. Macro-level approaches selectively process informative regions by leveraging clustering algorithms~\cite{clusdet_2019_iccv, crenet_2020_eccvw, pan_2022_cvpr, spdet_2023_icme} and density maps~\cite{dmnet_2020_cvprw, saccadedet_2024_ecmlpkdd, gigadet_2022_ncp} to isolate patches rich in small objects, or by utilizing non-uniform sampling~\cite{fovea_2021_iccv, salisa_2022_eccv, lzu_2023_cvpr, zoomdet_2026_isprs} to elastically compress backgrounds. Alternatively, micro-level architectures attempt to optimize efficiency by dynamically routing computation exclusively to informative tokens~\cite{sparseformer_2024_mm, gigamoe_2026_aaai, ceasc_2023_cvpr}. 

Although these efficiency-oriented designs substantially reduce computational overhead, they often sacrifice contextual information in UHR imagery, which is crucial for the semantic understanding of small objects. To address this issue, we propose a Global–Local Decoupled Decoder that reintroduces global context at the decoding stage, thereby enhancing the representation of small objects.

% While these efficiency-oriented designs significantly alleviate computational burdens, directly applying them to massive scenes remains non-trivial. Macro-level downsampling destroys small object features, while patch-based processing inherently severs continuous global context. Concurrently, micro-level routing risks prematurely discarding weak small objects as background noise. Consequently, a persistent dilemma remains: preserving high-resolution details for small objects often sacrifices macroscopic scene awareness, whereas maintaining global context typically leads to the loss of local details.

% However, directly applying these methods to massive scenes remains non-trivial. Macro-level downsampling destroys small object features, while patch-based processing inherently severs continuous global context. Concurrently, micro-level routing risks prematurely discarding weak small objects as background noise. Consequently, a persistent dilemma remains: preserving high-resolution details for small objects often sacrifices macroscopic scene awareness, whereas maintaining global context typically leads to the loss of local details.

\begin{figure*}[t]
    \centering
    \includegraphics[width=1.0\linewidth]{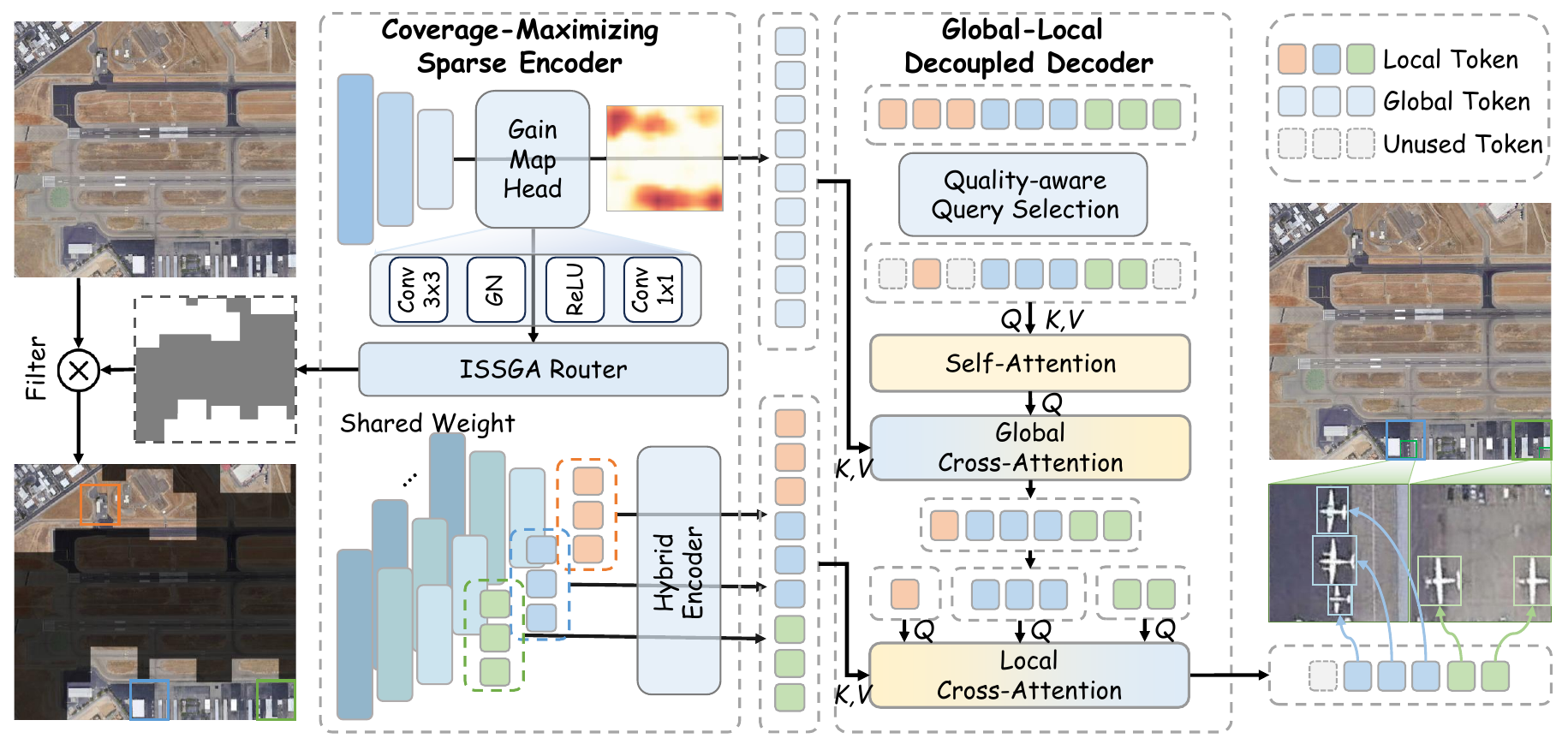}
    \caption{\textbf{Overall architecture of the proposed UHR-DETR.} The framework comprises two modules. (Left) The \textbf{Coverage-Maximizing Sparse Encoder} utilizes a lightweight backbone to generate global tokens and an object-aware Gain Map. Guided by this map, an Iterative Soft-Subtraction Greedy Algorithm (ISSGA) dynamically extracts high-value image patches, which are subsequently processed by the primary feature extractor to construct fine-grained local tokens. (Right) The \textbf{Global-Local Decoupled Decoder} hierarchically refines object queries. It sequentially performs global cross-attention to inject macroscopic spatial priors, followed by local cross-attention to accurately perceive microscopic details, yielding a unified and highly efficient end-to-end UHR perception pipeline.}
    \label{fig:architecture}
\end{figure*}

\section{Approach}
\label{methods}

\subsection{Overall Architecture}

% The overall architecture of the proposed UHR-DETR, illustrated in Fig.~\ref{fig:architecture}, is designed as an end-to-end perception pipeline comprising two principal modules: a \textbf{Coverage-Maximizing Sparse Encoder} and a \textbf{Global-Local Decoupled Decoder}. Unlike traditional architectures that uniformly process the entire image or isolate it into disconnected crops, UHR-DETR acts as a dynamic spatial router. It leverages a coarse global view to strategically sample high-resolution local patches, and subsequently employs a decoupled decoding mechanism to explicitly fuse macroscopic scene context with the fine-grained details of the selected regions, directly outputting global bounding boxes.

% revised
The overall architecture of the proposed UHR-DETR, illustrated in Fig.~\ref{fig:architecture}, is tailored for efficient object detection in ultra-high-resolution imagery. Unlike existing DETR-style frameworks that primarily target object detection under conventional resolutions, UHR-DETR is specifically designed to address the unique computational and contextual challenges posed by UHR inputs. The framework consists of two key components: \textbf{Coverage-Maximizing Sparse Encoder} and \textbf{Global–Local Decoupled Decoder}.
Coverage-Maximizing Sparse Encoder aims to efficiently encode UHR images by sparsely sampling informative regions under global guidance, thereby reducing computational overhead while preserving essential contextual coverage. Global–Local Decoupled Decoder is designed to integrate global scene context with fine-grained local features during decoding, enabling accurate localization and recognition of small objects.

Formally, given a massive UHR input image $\mathcal{I} \in \mathbb{R}^{H \times W \times 3}$, it is first downsampled via a max pooling layer. Then, a lightweight feature extractor (\textit{e.g.}, ResNet-18\cite{resnet_2016_cvpr}), denoted as $\Phi_{\text{light}}$, processes this downsampled input to extract global tokens $\mathcal{F}_{\text{global}} \in \mathbb{R}^{h \times w \times c}$, capturing the macroscopic topology of the scene:
\begin{equation}
    \mathcal{F}_{\text{global}} = \Phi_{\text{light}}\big(\operatorname{MaxPool}(\mathcal{I})\big).
\end{equation}
% $$ \mathcal{F}_{global} = \Phi_{light}\big(\operatorname{MaxPool}(\mathcal{I}, s)\big). $$

A Gain Map head subsequently acts on $\mathcal{F}_{\text{global}}$ to predict an object-aware Gain Map, which explicitly quantifies the potential object coverage of any given sub-region. Guided by this map, an iterative greedy algorithm strategically crops $K$ (default $K=40$) high-value image patches $\{\mathcal{P}_k\}_{k=1}^K$ from the original high-resolution image $\mathcal{I}$. These sparse but highly informative local regions are then fed into the primary feature extractor (we use the backbone and hybrid encoder of RT-DETR\cite{rt_detr_2024_cvpr} as our default feature extractor), parameterized as $\Phi_{p}$, to construct the high-resolution local tokens $\mathcal{F}_{\text{local}}$:
\begin{equation}
    \mathcal{F}_{\text{local}} = \bigcup_{k=1}^K \Phi_{p}(\mathcal{P}_k).
\end{equation}
% $$ \mathcal{F}_{local} = \bigcup_{k=1}^K \Phi_{p}(\mathcal{P}_k). $$

Finally, a set of object queries is processed through the Global-Local Decoupled Decoder. By sequentially performing Global Cross-Attention with $\mathcal{F}_{\text{global}}$ and Local Cross-Attention with $\mathcal{F}_{\text{local}}$, the queries seamlessly integrate macroscopic scene awareness with microscopic object details, ultimately predicting accurate bounding boxes in a unified global coordinate system.

\subsection{Coverage-Maximizing Sparse Encoder}

Processing the entire UHR image at its original resolution incurs prohibitive computational costs and severe memory bottlenecks, especially given that small objects are typically sparsely distributed across vast empty backgrounds. To resolve this inefficiency, the Coverage-Maximizing Sparse Encoder is designed to dynamically allocate finite computational resources to highly informative regions. %We formulate this selective spatial routing as a geometric set cover problem
To achieve this, we evaluate the perceptual value of each candidate sub-region across the vast UHR canvas. Guided by these regional value estimates, we then formulate the selective spatial routing as a geometric set cover problem, aiming to maximize the inclusion of small objects while minimizing spatial redundancy. Specifically, this module operates through a systematic two-step pipeline: it first performs \textit{global object distribution estimation} by predicting a dense Gain Map to quantify macroscopic object density, and subsequently executes \textit{dynamic sparse patch routing} via a tailored iterative greedy algorithm to strategically sample the optimal subset of high-resolution patches.

\subsubsection{Global Object Distribution Estimation}
To explicitly identify high-value regions without processing the entire UHR image at full resolution, this module is designed to estimate the spatial distribution of small objects. Taking the global tokens $\mathcal{F}_{\text{global}} \in \mathbb{R}^{h \times w \times c}$ as input, it produces a spatial Gain Map. The spatial dimensions $h \times w$ naturally divide the original UHR canvas into a dense grid of candidate patches. Each location $(x,y)$ on the Gain Map outputs a value score, representing the expected object coverage within a high-resolution patch of size $p_H \times p_W$ (set to $512 \times 512$ by default) centered at that coordinate.

Specifically, to robustly estimate the object quantity within each patch, we recast this unbounded dense regression into a classification-based expectation task. The Gain Map Head projects the input features $\mathcal{F}_{\text{global}}$ into a logit tensor $Z \in \mathbb{R}^{h \times w \times (M+1)}$, where $M$ is a predefined hyperparameter representing the discrete bin limit. Let $Z_{x,y,b}$ denote the predicted logit for the discrete bin $b \in \{0, 1, \dots, M\}$ at spatial location $(x,y)$. The final continuous predicted score, denoted by $\mathcal{G}_{(x,y)}$, is computed by mapping the discrete probability distribution to the continuous coverage scale~\cite{gfl_2020_nips}:
\begin{equation}
    \mathcal{G}_{(x,y)} = \sum_{b=0}^{M} b^2 \cdot \frac{\exp(Z_{x,y,b})}{\sum_{j=0}^{M} \exp(Z_{x,y,j})}.
\end{equation}
% $$ \mathcal{G}_{(x,y)} = \sum_{b=0}^{M} b^2 \cdot \frac{\exp(Z_{x,y,b})}{\sum_{j=0}^{M} \exp(Z_{x,y,j})}. $$

During training, this predicted score is supervised by the actual object coverage. The bounded ground-truth target, denoted as $\mathcal{G}^{GT}_{(x,y)}$, is defined based on the Intersection over Foreground (IoF). To prevent regions with extremely dense object clusters from dominating the learning gradients and to ensure optimization stability, we directly cap the raw IoF sum at a maximum threshold of $M^2$:
% (with $M$ set to 6 by default):
\begin{equation}
    \mathcal{G}^{GT}_{(x,y)} = \min\left(\sum_{i=1}^{N} \frac{\text{Area}(\mathcal{P}_{(x,y)} \cap B_i)}{\text{Area}(B_i)}, M^2\right),
\end{equation}
% $$ \mathcal{G}^{GT}_{(x,y)} = \min\left(\sum_{i=1}^{N} \frac{\text{Area}(\mathcal{P}_{(x,y)} \cap B_i)}{\text{Area}(B_i)}, M^2\right), $$
where $\text{Area}(B_i)$ denotes the spatial area of the $i$-th ground-truth bounding box. We then apply the Distribution Focal Loss~\cite{gfl_2020_nips} to supervise the discrete probability distribution against this continuous target $\mathcal{G}^{GT}_{(x,y)}$.

Furthermore, to explicitly sharpen the predicted local maxima around object clusters, we design an auxiliary objective termed the \textbf{Local Peak Margin Loss}. Denoting $\Omega$ as the set of ground-truth peak locations and $\Delta p$ as the 4-neighborhood shifts, the loss $\mathcal{L}_{LPM}$ penalizes any neighboring prediction $\mathcal{G}_{p+\Delta p}$ that encroaches within a predefined margin $\Delta$ of the peak value $\mathcal{G}_{p}$:
\begin{equation}
    \mathcal{L}_{LPM} = \frac{1}{|\Omega|} \sum_{p \in \Omega} \max\left(0, \mathcal{G}_{p+\Delta p} + \Delta - \mathcal{G}_{p}\right).
\end{equation}
% $$ \mathcal{L}_{LPM} = \frac{1}{|\Omega|} \sum_{p \in \Omega} \max\left(0, \mathcal{G}_{p+\Delta p} + \Delta - \mathcal{G}_{p}\right). $$

\subsubsection{Dynamic Sparse Patch Routing}

The core objective of this module is to dynamically allocate limited computational resources to the most valuable regions, effectively bypassing uninformative backgrounds. Taking the predicted continuous Gain Map $\mathcal{G}$ as input, this process outputs a sparse set of $K$ high-resolution image patches that collectively encapsulate the maximum number of objects.

To achieve this, we formulate this spatial routing as a geometric Maximum Coverage Problem, a classic formulation in combinatorial optimization proposed to identify a limited number of subsets that collectively cover the largest possible number of elements. Let the universe of elements $\mathcal{U}$ represent all objects of interest within the UHR image. Each spatial location $(x, y)$ defines a candidate subset $S_{(x,y)} \subseteq \mathcal{U}$, containing the objects physically covered by a local patch $\mathcal{P}_{(x,y)}$ centered at that coordinate. Our objective is to find a set of spatial coordinates, denoted as $\mathcal{C}$, that maximizes the total number of unique covered objects given a specific patch budget $K$:
\begin{equation}
    \max_{\mathcal{C}} \left| \bigcup_{(x,y) \in \mathcal{C}} S_{(x,y)} \right| \quad \text{s.t.} \quad |\mathcal{C}| = K.
\end{equation}
% $$\max_{\mathcal{C}} \left| \bigcup_{(x,y) \in \mathcal{C}} S_{(x,y)} \right| \quad \text{s.t.} \quad |\mathcal{C}| = K.$$

The predicted gain $\mathcal{G}_{(x,y)}$ acts as a continuous surrogate for the subset cardinality $|S_{(x,y)}|$, estimating the theoretical value of extracting fine-grained features from this region. Since the Maximum Coverage Problem is inherently NP-hard (i.e., finding the absolute optimal subset is computationally intractable in polynomial time), we approximate the optimal solution using a standard greedy approach. This method iteratively selects the highest-value subset and subsequently removes the covered elements from all remaining candidates.

% The predicted gain ${\mathcal{G}}_{(x,y)}$ acts as a continuous surrogate for the subset cardinality $|S_{(x,y)}|$, estimating the theoretical value of extracting fine-grained features from this region. Since the Maximum Coverage Problem is inherently NP-hard, we approximate the optimal solution using a standard greedy approach. This method iteratively selects the highest-value subset and subsequently removes the covered elements from all remaining candidates.

Since explicit target coordinates are unknown prior to fine-grained detection, we approximate this element removal process using an Iterative Soft-Subtraction Greedy Algorithm, detailed in Algorithm \ref{alg:greedy}.

In each iteration, the algorithm identifies the center coordinate $(x_k, y_k)$ exhibiting the peak gain $v_k$. Rigidly zeroing out this selected region would strictly forbid patch overlapping, leading to sub-optimal coverage for dense target clusters spanning across patch boundaries. Instead, we implement a soft-subtraction mechanism to permit necessary intersections. We construct a 2D linear decay overlap kernel $\mathcal{K}$ to mathematically quantify the spatial overlapping penalty. For any spatial offset $(\Delta x, \Delta y)$ relative to the selected center, the kernel is formally defined as:
\begin{equation}
    \mathcal{K}(\Delta x, \Delta y) = \max\left(0, 1 - \frac{|\Delta x|}{p_w}\right) \cdot \max\left(0, 1 - \frac{|\Delta y|}{p_h}\right),
\end{equation}
% $$\mathcal{K}(\Delta x, \Delta y) = \max\left(0, 1 - \frac{|\Delta x|}{p_w}\right) \cdot \max\left(0, 1 - \frac{|\Delta y|}{p_h}\right),$$
where $p_w$ and $p_h$ denote the patch dimensions projected onto the spatial grid of the Gain Map. Upon selecting the peak $v_k$, the remaining Gain Map is softly discounted to simulate the removal of covered elements:
\begin{equation}
    \mathcal{G}_{(x,y)} \leftarrow \max\left(0, \mathcal{G}_{(x,y)} - v_k \cdot \mathcal{K}(x - x_k, y - y_k)\right).
\end{equation}
% $$\mathcal{G}_{(x,y)} \leftarrow \max\left(0, \mathcal{G}_{(x,y)} - v_k \cdot \mathcal{K}(x - x_k, y - y_k)\right).$$

This iterative suppression explicitly penalizes severe overlaps while permitting slight, necessary intersections, ensuring the selected $K$ patches are both diverse and densely packed around target clusters. The cropped patches are then processed by the primary encoder to construct the fine-grained $\mathcal{F}_{\text{local}}$.

\begin{algorithm}[t]
% \small
\caption{Iterative Soft-Subtraction Greedy Algorithm}
\label{alg:greedy}
\begin{algorithmic}[1]
\Require Gain Map $\mathcal{G} \!\in\! \mathbb{R}^{h \times w}$,
  Patch Size $p_h \!\times\! p_w$, Budget $K$
\Ensure Coordinates $\mathcal{C}$, Scores $\mathcal{S}$

\State $\mathcal{C} \gets \emptyset$;\;
       $\mathcal{S} \gets \emptyset$

\Statex $\triangleright$
  \textit{Build linear decay overlap kernel}
\For{$\Delta x \!\in\! [\text{-}p_w,\, p_w]$,\;
     $\Delta y \!\in\! [\text{-}p_h,\, p_h]$}
  \State $\mathcal{K}(\Delta x, \Delta y) \gets
    \Big(1 \!-\! \tfrac{|\Delta x|}{p_w}\Big)^{\!+}
    \!\!\cdot
    \Big(1 \!-\! \tfrac{|\Delta y|}{p_h}\Big)^{\!+}$
\EndFor

\Statex $\triangleright$
  \textit{Greedy selection with soft suppression}
\For{$k = 1$ \textbf{to} $K$}
  \State $(x_k,\, y_k) \gets
    \arg\max_{x,y}\, \mathcal{G}_{(x,\, y)}$
  \State $v_k \gets
    {\mathcal{G}}_{(x_k,\, y_k)}$
  \State $\mathcal{C} \!\gets\! \mathcal{C} \cup \{(x_k, y_k)\}$;\;
         $\mathcal{S} \!\gets\! \mathcal{S} \cup \{v_k\}$
  \If{$v_k > 0$}
    \For{\textbf{each}
      $(x, y) \in \mathcal{N}_{p_w, p_h}(x_k, y_k)$}
      \State $\mathcal{G}_{(x,y)} \gets
        \bigl({\mathcal{G}}_{(x,y)}$
      % \Statex \hspace{4.2em}
        $-\; v_k \!\cdot\!
        \mathcal{K}(x \!-\! x_k,\;
                     y \!-\! y_k)\bigr)^{\!+}$
    \EndFor
    \State $\mathcal{G}_{(x_k,\, y_k)}
      \gets 0$
  \EndIf
\EndFor

\State \Return $\mathcal{C}$, $\mathcal{S}$
\end{algorithmic}
\vspace{2pt}
\footnotesize
$(z)^{+}\!=\!\max(z,0)$;\;

$\mathcal{N}_{p_w,p_h}(x_k,y_k) \!=\!
 \{(x,y): |x\!-\!x_k|\!\leq\! p_w,\,
           |y\!-\!y_k|\!\leq\! p_h\}
 \cap\, [0,w)\!\times\![0,h)$.
 % $(z)^{+}\!=\!\max(z,0)$.
\end{algorithm}

\subsection{Global-Local Decoupled Decoder}

To resolve the severe context loss inherent in isolated patch inference, our Global-Local Decoupled Decoder divides the semantic interaction into two hierarchical stages: \textit{Global Macro-Interaction} and \textit{Local Micro-Interaction}. This decoupled design seamlessly bridges macroscopic scene awareness with microscopic object details.

% Leveraging the macroscopic global tokens $\mathcal{F}_{global}$ and the microscopic local tokens $\mathcal{F}_{local}$ generated by the Coverage-Maximizing Sparse Encoder, we first initialize a set of object queries $Q$. The number of $Q$ is dynamically determined proportional to the spatial sum of the predicted Gain Map, implicitly adapting to the actual scene density. Within each decoder layer, these queries undergo a standard Self-Attention, followed by a decoupled dual cross-attention pipeline. This pipeline systematically aggregates features from both spatial scales through two distinct interaction phases.
Leveraging the macroscopic global tokens $\mathcal{F}_{\text{global}}$ and the microscopic local tokens $\mathcal{F}_{\text{local}}$ generated by the Coverage-Maximizing Sparse Encoder, we first initialize a set of object queries $Q$. The number of $Q$ is dynamically determined proportional to the spatial sum of the predicted Gain Map, implicitly adapting to the actual scene density. Specifically, each decoder layer is structured as a systematic three-step interaction pipeline: it begins with a self-attention module to capture inter-object dependencies, followed by a global cross-attention module to mine the correlations between objects and the global context, and concludes with a local cross-attention module to associate the objects with local fine-grained features.

% \textit{Global Macro-Interaction:} To avoid being blindly trapped in isolated patches and to acquire essential holistic context, the object queries perform Multi-Head Cross-Attention (MHCA)~\cite{attention_2017_nips} with the global tokens $\mathcal{F}_{global}$. Following DAB-DETR~\cite{dab_detr_2022_iclr}, we formulate the queries as dynamic anchor boxes to explicitly parameterize their spatial priors. Let $A_{global} \in [0,1]^4$ denote the queries' anchor boxes (comprising center coordinates, width, and height) in the absolute normalized coordinate space of the original UHR image. By applying a sinusoidal positional encoding function $\text{PE}(\cdot)$, the anchor boxes are spatially aligned with the global feature map. The interaction is thus formulated as:
% $$ Q_{global} = \text{MHCA}(Q + \text{PE}(A_{global}), \mathcal{F}_{global} + \text{PE}(G_{global}), \mathcal{F}_{global}), $$
% where $G_{global}$ represents the standard spatial grid coordinates of the global tokens. This step injects crucial macroscopic spatial priors, effectively circumventing the contextual limitations inherent to disconnected patch inference.

\textit{Global Macro-Interaction:} To avoid being blindly trapped in isolated patches and to acquire essential holistic context, the object queries perform Multi-Head Cross-Attention (MHCA)~\cite{attention_2017_nips} with the global tokens $\mathcal{F}_{\text{global}}$. Following DAB-DETR~\cite{dabdetr_2022_iclr}, we formulate the queries as dynamic anchor boxes to explicitly parameterize their spatial priors. Let $A = (cx, cy, w, h) \in [0,1]^4$ denote the queries' anchor boxes in the absolute normalized coordinate space. By applying a sinusoidal positional encoding function $\text{PE}(\cdot)$, the anchor boxes are spatially aligned with the global feature map. The interaction is thus formulated as:
\begin{equation}
    Q_{\text{global}} = \text{MHCA}(Q + \text{PE}(A), \mathcal{F}_{\text{global}} + \text{PE}(X), \mathcal{F}_{\text{global}}),
\end{equation}
% $$ Q_{global} = \text{MHCA}(Q + \text{PE}(A), \mathcal{F}_{global} + \text{PE}(X), \mathcal{F}_{global}), $$
where $X$ represents the standard spatial grid coordinates of the global tokens. This step injects crucial macroscopic spatial priors, effectively circumventing the contextual limitations inherent to disconnected patch inference.

\textit{Local Micro-Interaction:} Once imbued with global awareness, the global queries $Q_{\text{global}}$ are dynamically routed and partitioned into subsets $\{Q^{(k)}\}_{k=1}^K$ corresponding to their assigned spatial patches. To perform precise micro-level feature extraction, the anchor boxes of each query subset must be projected from the global normalized coordinate system to the patch-local coordinate system. For a local patch $\mathcal{P}_k$ defined by its pixel boundaries $(x_1, y_1, x_2, y_2)$ within an original image of width $W$ and height $H$, the local continuous center coordinate $cx_{\text{local}}$ is computed from the global center $cx$ as:
\begin{equation}
    cx_{\text{local}} = \frac{cx \cdot W - x_1}{x_2 - x_1}, 
\end{equation}
% $$ cx_{local} = \frac{cx \cdot W - x_1}{x_2 - x_1}, $$
and similarly for $cy_{\text{local}}$. Alongside the proportional scaling of width and height, this transformation yields the localized anchor boxes $A^{(k)}_{\text{local}}$. Subsequently, a Multi-Scale Deformable Attention (MSDA)~\cite{deformable_detr_2021_iclr} mechanism is applied independently within each patch group. This restricts the dense key-value interactions between the grouped queries $Q^{(k)}$ and their corresponding high-resolution local tokens $\mathcal{F}^{(k)}_{\text{local}}$:
\begin{equation}
    \widetilde{Q}^{(k)} = \text{MSDA}(Q^{(k)}, A^{(k)}_{\text{local}}, \mathcal{F}^{(k)}_{\text{local}}).
\end{equation}
% $$ \widetilde{Q}^{(k)} = \text{MSDA}(Q^{(k)}, A^{(k)}_{local}, \mathcal{F}^{(k)}_{local}). $$

After the localized interaction, the refined query subsets $\{ \widetilde{Q}^{(k)} \}_{k=1}^K$ are scattered back to reconstruct the unified query tensor, which is then processed by a Feed-Forward Network. By decoupling the interactions into global and local hierarchical stages, the model achieves simultaneous perception of both macroscopic scene context and microscopic object details, ensuring highly precise bounding box refinement.

\subsection{Optimization Objectives}
UHR-DETR is trained in an end-to-end manner. The overall loss function is formulated as a linear combination of the Gain Map generation losses and the set prediction losses:
\begin{equation}
    \mathcal{L}_{total} = \lambda_{map} \mathcal{L}_{map} + \lambda_{LPM} \mathcal{L}_{LPM} + \mathcal{L}_{det},
\end{equation}
% $$ \mathcal{L}_{total} = \lambda_{map} \mathcal{L}_{map} + \lambda_{LPM} \mathcal{L}_{LPM} + \mathcal{L}_{det}, $$
where $\mathcal{L}_{map}$ denotes the Distribution Focal Loss for the Gain Map, and $\mathcal{L}_{LPM}$ represents the Local Peak Margin Loss. The detection loss $\mathcal{L}_{det}$ is computed based on the standard bipartite matching protocol utilizing the Hungarian algorithm, comprising Varifocal Loss for classification alongside $L_1$ and GIoU losses for bounding box regression.

\begin{table*}[!t]
\setlength{\tabcolsep}{2pt}
\renewcommand{\arraystretch}{1.2}
\caption{Detection results on the STAR dataset. $\dag$ denotes downsampling to $1024 \times 1024$. $*$ indicates training from scratch for 36 epochs following \cite{sparseformer_2024_mm}.}
\label{tab:main_res_star}
\centering
\resizebox{0.99\textwidth}{!}{

\begin{tabular}{r | r|c|cccccccc|cccc|cc}
\toprule
ID & Method       & Backbone  & Car   & Boat  & Tank  & Truck & BB   & Crane &Airplane& LT    & mAP   &$AP_S$ &$AP_M$ &$AP_L$ & Latency(s)   & Mem.(GB) \\ 
\midrule
\multicolumn{17}{l}{\textit{Exhaustive Sliding-Window Baselines (SW):}} \\
\midrule
\#1 & Faster-RCNN\cite{faster_rcnn_2016_tpami}& ResNet-50 & 5.3   & 11.6  & 39.0  & 10.7  & 51.0  & 17.6  & 80.4  & 50.9  & 33.3  & 14.8  & 34.6  & 38.8  & 4.839     &  4.85 \\
\#2 & FCOS\cite{fcos_2019_iccv}& ResNet-50 & 4.6   & 6.5   & 35.4  & 7.3   & 50.2  & 14.0  & 79.0  & 38.6  & 29.5  & 12.5  & 31.1  & 36.6  & 3.384     &  2.27 \\
\#3 & YOLOX-x\cite{yolox_2021_arxiv} & CSPDarknet& 5.1   & 1.6   & 25.3  & 7.7   & 18.4  & 7.3   & 72.8  & 25.5  & 20.4  & 7.8   & 21.7  & 27.2  & 5.244     &  3.62   \\
\#4 & RT-DETR\cite{rt_detr_2024_cvpr}& ResNet-50 & 4.7   & 10.9  & 30.9  & 7.0   & 62.3  & 16.2  & 80.5  & 44.0  & 32.1  & 14.0  & 33.1  & 37.7  & 3.068     &  2.32 \\
\midrule
\multicolumn{17}{l}{\textit{Global Downsampling Approaches:}} \\
\midrule
\#5 & RT-DETR\dag\cite{rt_detr_2024_cvpr}& ResNet-50 & 1.4   & 1.8   & 15.8  & 8.9   & 6.8   & 14.3  & 39.2  & 21.8  & 13.8  & 0.6   & 9.5   & 31.4  & 0.036     & 0.46  \\
\#6 & Fovea\dag\cite{fovea_2021_iccv}& ResNet-50 & 0.0   & 0.0   & 12.6  & 4.3   & 3.3   & 8.2   & 35.2  & 19.3  & 10.3  & 0.2   & 6.4   & 24.8  & 1.560     & 1.97    \\
\midrule
\multicolumn{17}{l}{\textit{Micro-level Dynamic Routing \& SW:}} \\
\midrule
% SparseFormer(3x)&SparseFormer& 5.8  & 1.8   & 24.3  & 5.9   & 47.4  & 9.3   & 77.1  & 20.6  & 24.0  & 7.2   & 26.4  & 28.6  & 7.450    &  4.30 \\
\#7 & SparseFormer*\cite{sparseformer_2024_mm} & SparseNet &  4.9  & 0.6   & 25.2  & 5.8   & 44.8  & 12.7  & 74.2  & 17.3  & 23.2  & 7.6   & 23.8  & 30.4  & 4.146     & 2.51  \\
\#8 & CEASC\cite{ceasc_2023_cvpr}& ResNet-50 &  5.6  & 8.5   & 31.8  & 10.4  & 51.5  & 16.2  & 78.2  & 43.1  & 30.7  & 14.7  & 31.4  & 36.7  & 3.820     & 5.00 \\  
\midrule
\multicolumn{17}{l}{\textit{Macro-level Patch Selection:}} \\
\midrule
\#9 & SPDet\cite{spdet_2023_icme}& CSPDarknet&  1.1  & 0.1   & 22.9  & 1.9   & 15.6  & 4.3   & 62.2  & 16.6  & 15.6  & 0.8   & 15.1  & 26.9  & 2.580     &  0.75    \\
\#10 & GigaDet\cite{gigadet_2022_ncp}& ResNet-50 &  1.1  & 1.2   & 20.8  & 3.7   & 23.5  & 1.3   & 50.7  & 11.5  & 14.2  & 4.6   & 15.2  & 17.8  & 0.345     &  3.12     \\  
\midrule
\rowcolor{gray!20}
\#11 & UHR-DETR(ours)  & ResNet-50 & 14.5  & 10.4  & 34.6  & 33.9  & 52.4  & 20.2  & 70.1  & 43.2  & 34.9  & 15.0  & 36.4  & 40.5  & 0.357     &  3.13 \\
\bottomrule
\end{tabular}
}
\end{table*}

\section{Experiments}
\label{experiments}

% In this section, we comprehensively evaluate the proposed UHR-DETR. We first introduce the benchmark datasets and our specific implementation details. Subsequently, we compare our approach with state-of-the-art detectors and conduct thorough ablation studies to validate the effectiveness of our core architectural designs.

\subsection{Datasets}
We conduct extensive experiments on two challenging large-scale datasets featuring massive image resolutions and densely packed small objects: STAR \cite{star_2025_tpami} and SODA-A \cite{soda_2023_tpami}.

\textbf{STAR}\cite{star_2025_tpami}, originally proposed for scene graph generation, provides a challenging benchmark for object detection in satellite imagery. The dataset features high spatial resolutions ($0.15$m to $1$m), with original image dimensions spanning from $512 \times 768$ to $27,860 \times 31,096$ pixels. Overall, it contains over $210,000$ instances annotated across $48$ fine-grained categories.

\textbf{SODA-A}\cite{soda_2023_tpami} is a large-scale benchmark exclusively tailored for multi-category small object detection in aerial scenarios. It comprises $2,513$ high-resolution aerial images encompassing over $872,000$ exhaustively annotated instances across $9$ distinct classes.

\textbf{Data Preprocessing.} To construct standardized evaluation benchmarks for UHR small object detection, we apply specific preprocessing strategies to both datasets. For STAR, we first select the $8$ categories exhibiting the smallest average object sizes across the entire dataset: \textit{car}, \textit{boat}, \textit{tank}, \textit{truck}, \textit{boarding\_bridge}, \textit{crane}, \textit{airplane}, and \textit{lattice\_tower}, unifying all annotations into the standard $(x, y, w, h)$ representation. To formulate a specialized subset, we filter the dataset to retain only images that contain instances of these selected categories and possess an original resolution exceeding 6K. Finally, we employ a sliding window and zero-padding strategy to standardize the retained images to a fixed resolution of $8192 \times 8192$ pixels~\cite{geollava8k_2025_nips}. For SODA-A, we retain all original object categories. To simulate UHR conditions, we apply a $2\times$ upsampling to the original images. Since the long edge of images in SODA-A is typically 4800 pixels, this scaling extends the maximum dimension to 9600 pixels. To accommodate the varying short edges, we then apply zero-padding to standardize all images to a fixed resolution of $9600 \times 9600$ pixels.

\subsection{Evaluation Metrics}
To evaluate detection performance, we report the mean Average Precision (mAP) alongside $\text{AP}_S$, $\text{AP}_M$, and $\text{AP}_L$. All precision metrics are calculated at a fixed Intersection over Union (IoU) threshold of $0.5$. Additionally, we provide the average precision for each individual category for fine-grained analysis. To assess computational efficiency, we measure Latency, defined as the average inference time per UHR image, and report the GPU memory footprint (Mem.) during inference for reference.

% \subsection{Implementation Details}
% We implement UHR-DETR using the MMDetection~\cite{mmdet_2019_arxiv} framework and conduct all experiments on a single NVIDIA RTX 3090 GPU with a batch size of 1. The backbone for extracting local features defaults to ResNet-50~\cite{resnet_2016_cvpr}, while a lightweight ResNet-18~\cite{resnet_2016_cvpr} is utilized for extracting global features. Both backbones are initialized with ImageNet~\cite{imagenet_2015_ijcv} pre-trained weights. During the patch selection phase, the local patch size is set to $512 \times 512$ pixels. The number of selected patches $K$ defaults to 40 for STAR and 20 for SODA-A. The discrete bin limit $M$ is set to 6 based on dataset statistics, as over 99\% of sub-regions exhibit a ground-truth coverage below 36 ($M^2$). For the decoupled decoder, the number of dynamically initialized object queries is bounded between 300 and 3000. The entire network is trained end-to-end for 12 epochs using the AdamW~\cite{adamw_2019_iclr} optimizer with an initial learning rate of $1 \times 10^{-4}$ and a weight decay of $1 \times 10^{-4}$. To preserve pre-trained representations, the learning rate for both backbones is scaled by a factor of 0.1. We apply a linear warmup strategy for the first 500 iterations and decay the overall learning rate by a factor of 10 at the 8th and 11th epochs. Finally, gradient clipping with a maximum norm of 0.1 is adopted to stabilize the training process.

\subsection{Implementation Details}
We implement UHR-DETR using the MMDetection~\cite{mmdet_2019_arxiv} framework and conduct all experiments on a single NVIDIA RTX 3090 GPU with a batch size of 1. The backbone for extracting local features defaults to ResNet-50~\cite{resnet_2016_cvpr}, while a lightweight ResNet-18~\cite{resnet_2016_cvpr} is utilized for extracting global features. Both backbones are initialized with ImageNet~\cite{imagenet_2015_ijcv} pre-trained weights. During the patch selection phase, the local patch size is set to $512 \times 512$ pixels. The number of selected patches $K$ defaults to 40 for STAR, while it is set to 80 for SODA-A to accommodate the more uniform object distribution within its simulated UHR scenes. The discrete bin limit $M$ is set to 6 based on dataset statistics, as over 99\% of sub-regions exhibit a ground-truth coverage below 36 ($M^2$). For the decoupled decoder, the number of dynamically initialized object queries is bounded between 300 and 3000. The entire network is trained end-to-end for 12 epochs using the AdamW~\cite{adamw_2019_iclr} optimizer with an initial learning rate of $1 \times 10^{-4}$ and a weight decay of $1 \times 10^{-4}$. To preserve pre-trained representations, the learning rate for both backbones is scaled by a factor of 0.1. We apply a linear warmup strategy for the first 500 iterations and decay the overall learning rate by a factor of 10 at the 8th and 11th epochs. Finally, gradient clipping with a maximum norm of 0.1 is adopted to stabilize the training process.

\subsection{Main Results}
To demonstrate the superiority of UHR-DETR, we conduct comprehensive comparisons with various state-of-the-art detectors on the STAR and SODA-A benchmarks. As detailed in Table~\ref{tab:main_res_star} and Table~\ref{tab:main_res_sodaa}, the competing methods are systematically categorized into four paradigms: Exhaustive Sliding-Window Baselines, Global Downsampling Approaches, Micro-level Dynamic Routing (with Sliding Windows), and Macro-level Patch Selection.

\begin{table*}[t!]
\normalsize
\setlength{\tabcolsep}{2pt}
\renewcommand{\arraystretch}{1.4}
\caption{Detection results on the SODA-A dataset. $\dag$ denotes downsampling to $1024 \times 1024$. $*$ indicates training from scratch for 36 epochs following \cite{sparseformer_2024_mm}.}
\label{tab:main_res_sodaa}
\centering
\resizebox{0.99\textwidth}{!}{

\begin{tabular}{r|r|c|ccccccccc|cccc|cc}
\toprule
ID & Method          & Backbone  & AI  & HE & SV    & LV    & SH  & CO & ST    &SP     & WI  & mAP   &$AP_S$ &$AP_M$ &$AP_L$ & Latency(s)& Mem.(GB) \\ 
\midrule
\multicolumn{18}{l}{\textit{Global Downsampling Approaches:}} \\
\midrule
\#1 & RT-DETR$\dag$\cite{rt_detr_2024_cvpr}& ResNet-50 &  29.0     &  23.4     &  2.3  & 4.8  & 8.1  &  19.0    &  11.1 & 48.0 & 19.0     & 18.3 & 1.2 & 21.0  &  34.3  & 0.036     & 0.46    \\   % mAR=53.8
\#2 & Fovea$\dag$\cite{fovea_2021_iccv}& ResNet-50 &  5.0     &  0.0    &   0.0  &  2.5 & 1.0 & 1.0    &  3.5 &  2.8 &   1.5  & 1.9  & 0.3   & 2.0  & 4.6      & 2.100     & 2.53      \\
\midrule
\multicolumn{18}{l}{\textit{Micro-level Dynamic Routing \& SW:}} \\
\midrule
\#3 & SparseFormer$*$\cite{sparseformer_2024_mm}& SparseNet &   80.0   &   62.6  & 47.1 &  37.3 &  61.4 &   46.8   &  63.3 & 83.1  & 40.6   &  58.0 & 24.0  &  68.0 &  53.6  &  4.802    &  2.51     \\
\#4 & CEASC\cite{ceasc_2023_cvpr}& ResNet-50 &  81.6    &  61.4     & 45.7  & 37.6  &  63.6 &  43.0   & 66.5 & 80.2 &  56.1    & 59.5 & 25.7  &  69.7 &   48.3   & 4.467     & 5.38      \\   %mAR=77.3
\midrule
\multicolumn{18}{l}{\textit{Macro-level Patch Selection:}} \\
\midrule
\#5 & SPDet\cite{spdet_2023_icme}& CSPDarknet&  64.5   &  8.2    &  31.9 & 20.9 & 42.8  &  38.8   & 55.1 & 70.8 &  18.0   & 39.0  & 13.1  & 48.5  &  39.3     & 1.957    &  0.75     \\
\#6 & GigaDet\cite{gigadet_2022_ncp}& ResNet-50 & 68.8      & 16.1     & 45.9  & 10.5  & 66.5  & 31.2      & 45.0  & 21.4   & 13.6     & 35.5  & 23.5  & 40.2  & 25.6    & 0.588     & 6.72      \\
\midrule
\rowcolor{gray!20}
\#7 & UHR-DETR(ours)& ResNet-50 & 83.8     & 53.4      & 45.9  & 20.8  & 56.8  & 36.3      & 69.4  & 51.4  & 61.9      & 53.3  & 30.9  & 59.4  & 40.8    & 0.647     & 5.57     \\
\bottomrule
\end{tabular}
}
\end{table*}

\textbf{Results on the STAR Dataset.} As presented in Table~\ref{tab:main_res_star}, traditional sliding-window baselines achieve considerable accuracy but incur prohibitive computational costs. Uniform downsampling (\#5) severely degrades small object detection. Although non-uniform downsampling (\#6) attempts to preserve fine-grained features, it still struggles to perceive small objects in UHR images, even resulting in performance regression. For similar reasons, dynamic routing and patch selection approaches (\#7) fail to strike an optimal balance between performance and efficiency. In contrast, our method (\#11) reduces the inference latency to 0.357s through the sparse extraction of fine-grained features, achieving nearly a 10$\times$ speedup compared to the sliding-window baselines. Furthermore, operating within a streamlined, fully end-to-end inference framework, our approach boosts the overall mAP to 34.9\% and the small object accuracy ($\text{AP}_S$) to 15.0\%. Particularly, for typical small objects like \textit{car}, our method achieves a precision of 14.5\%, significantly outperforming all other baselines.
%Notably, for inherently miniature and densely distributed categories such as \textit{car}, our method demonstrates overwhelming superiority, achieving a remarkable precision of 14.5\%, which substantially outperforms all other comparative baselines.

% \textbf{Results on the SODA-A Dataset.} To further validate the effectiveness of our method, we conduct experiments on the SODA-A dataset, where images are standardized to a 3K resolution. As shown in Table~\ref{tab:main_res_sodaa}, UHR-DETR achieves a competitive mAP of 50.5\% with an inference latency of 0.153s and a memory footprint of 1.41 GB. These results demonstrate that our approach maintains strong detection performance and high computational efficiency across different resolution scales.

% \textbf{Results on the SODA-A Dataset.} To further validate the effectiveness of our method, we evaluate UHR-DETR on the SODA-A dataset under simulated $9600 \times 9600$ conditions. As shown in Table~\ref{tab:main_res_sodaa}, compared to sliding-window methods (\#3 and \#4), our approach achieves a competitive mAP of 53.3\% while delivering a significantly faster inference latency of 0.647s. Simultaneously, it yields the highest small object accuracy ($AP_S$ of 30.9\%) among all evaluated baselines.

\textbf{Results on the SODA-A Dataset.} To further validate the effectiveness of our method, we evaluate UHR-DETR on the SODA-A dataset under simulated $9600 \times 9600$ conditions. As shown in Table~\ref{tab:main_res_sodaa}, compared to sliding-window methods (\#3 and \#4), our approach achieves a competitive mAP of 53.3\% while delivering a significantly faster inference latency of 0.647s. Simultaneously, it yields the highest small object accuracy ($AP_S$ of 30.9\%) among all evaluated baselines. Even when facing more challenging conditions in this simulated scenario, our method successfully suppresses background redundancy. This indicates that the proposed sparse perception paradigm remains robust across varying scene complexities.
However, despite achieving strong performance, the margin of improvement on SODA-A is smaller than that on the STAR dataset. This is primarily because SODA-A is constructed via uniform upsampling to obtain UHR images, resulting in a more homogeneous object distribution. In contrast, STAR consists of naturally acquired UHR imagery with more realistic observation distributions. Consequently, this difference poses greater challenges for efficient feature extraction in the Encoder. We further discuss this issue in detail in the Section~\ref{discussion}.

\subsection{Ablations}

\textbf{Effectiveness of Core Components.} As detailed in Table~\ref{tab:ablations}(a), our proposed components (ISSGA, LPM, and GCA) synergistically drive a substantial performance leap, ultimately boosting the baseline mAP from 27.4\% to 34.9\%. 
The baseline model, which employs rigid patch selection (NMS, similar to \cite{gigadet_2022_ncp}), yields 27.4\% mAP. Replacing this rigid strategy with ISSGA brings a 2.1\% mAP improvement (29.5\%) by effectively mitigating the truncation of dense object clusters. The subsequent addition of the LPM objective triggers a remarkable 4.3\% mAP surge (reaching 33.8\%). This highlights a crucial synergy: LPM explicitly sharpens local peaks in the Gain Map, providing optimal spatial guidance for the ISSGA routing. Finally, integrating Global Cross-Attention (GCA) elevates the overall mAP to 34.9\% and notably achieves a striking improvement for small objects ($\text{AP}_S$ surges from 12.9\% to 15.0\%). This confirms that injecting macroscopic context into local decoding is essential for recognizing small objects with weak isolated features.

% \textbf{Impact of Patch Quantity ($K$).} As shown in Table~\ref{tab:ablations}(b), our method demonstrates strong robustness across various patch budgets and allows for elastic, training-free adjustment of $K_{infer}$ during deployment to meet dynamic hardware constraints. Vertically, scaling up $K_{train}$ consistently raises the performance upper bound across all inference settings, as a larger budget exposes the network to a greater number of object instances and complex backgrounds. Horizontally, increasing $K_{infer}$ steadily improves mAP by processing more informative regions, but introduces a linear latency overhead. For instance, with $K_{train}=40$, pushing $K_{infer}$ from 40 to 60 yields a marginal mAP gain (34.9\% to 35.6\%) but substantially increases latency (from 357 ms to 479 ms). Balancing detection accuracy and computational efficiency, we adopt $K_{train}=40$ and $K_{infer}=40$ as our default configuration.
\textbf{Impact of Patch Quantity ($K$).} As shown in Table~\ref{tab:ablations}(b), our method demonstrates remarkable robustness to variations in patch budgets, enabling elastic, training-free adjustment of $K_{\text{infer}}$ during deployment. Vertically, scaling up $K_{\text{train}}$ consistently raises the performance upper bound, as a larger budget exposes the network to more diverse object instances. Horizontally, the model is highly insensitive to $K_{\text{infer}}$ variations. Increasing $K_{\text{infer}}$ steadily improves mAP without requiring strict alignment with $K_{\text{train}}$, though it naturally introduces a linear latency overhead. For instance, with $K_{\text{train}}=40$, pushing $K_{\text{infer}}$ from 40 to 60 yields a marginal mAP gain (34.9\% to 35.6\%) but substantially increases latency (357 ms to 479 ms). Balancing accuracy and efficiency, we adopt $K_{\text{train}}=40$ and $K_{\text{infer}}=40$ as our default configuration.

% \begin{table*}[t]\vspace{-3mm}
\begin{table*}[t]
\centering
% \caption{Ablations. We train on STAR \texttt{train set}, test on \texttt{test set}. The evaluation metric is mAP.
% ISSGA, LPM, and GCA denote the Iterative Soft-
% Subtraction Greedy Algorithm, Local Peak Margin Loss, and
% Global Cross-Attention in the decoder, respectively.
% Gray row means the default setting. }
\caption{Ablations. ISSGA, LPM, and GCA denote the Iterative Soft-
Subtraction Greedy Algorithm, Local Peak Margin Loss, and
Global Cross-Attention in the decoder, respectively.
Gray row means the default setting. }
% subfloat a
\subfloat[Ablation study on the core components of our method. \label{tab:ablation_components}]{
\tablestyle{9pt}{1.25}\begin{tabular}{ccc cccc}
        \toprule
        \multicolumn{3}{c}{Components} & \multicolumn{4}{c}{Metrics (\%)} \\
        \cmidrule(r){1-3} \cmidrule(l){4-7}
        ISSGA & LPM & GCA & mAP & $\text{AP}_S$ & $\text{AP}_M$ & $\text{AP}_L$ \\
        \midrule
        & & & 27.4 & 11.1 & 29.2 & 32.8 \\
        & & \checkmark & 28.3 & 11.9 & 30.0 & 32.8 \\
        \checkmark & & & 29.5 & 12.3& 31.3 & 33.2 \\
        \checkmark & \checkmark & & 33.8 & 12.9 & 35.4 & 40.2 \\
        \rowcolor{gray!20}
        \checkmark & \checkmark & \checkmark & 34.9 & 15.0 & 36.4 & 40.5 \\
        \bottomrule
    \end{tabular}}\hspace{3mm}
% subfloat b
\subfloat[Cross-evaluation of the number of selected patches during training ($K_{\text{train}}$) and inference ($K_{\text{infer}}$). The evaluation metric is mAP.
\label{tab:cross_k}]{
\tablestyle{11pt}{1.35}\begin{tabular}{c | ccccc}
    \toprule
    \multirow{2}{*}{$K_{\text{train}}$} & \multicolumn{5}{c}{Inference Patches ($K_{\text{infer}}$)} \\
    \cmidrule{2-6}
    & 20 & 30 & 40 & 50 & 60 \\
    \midrule
    20 & 29.9 & 31.4 & 31.8 & 31.8 & 31.7 \\
    30 & 30.6 & 32.3 & 33.0 & 32.9 & 33.0 \\
    40 & 31.7 & 33.7 & 34.9 & 35.3 & 35.6 \\
    \midrule
    % \rowcolor{gray!20}
    Latency (ms) & 238 & 298 & 357 & 420 & 479 \\
    \bottomrule
  \end{tabular}} \\
% subfloat c
\subfloat[Sensitivity analysis of the hyperparameter margin $\Delta$ in the Local Peak Margin Loss ($\mathcal{L}_{LPM}$). \label{tab:ablation_margin}]{
\tablestyle{9pt}{1.45}\begin{tabular}{c cccc}
        \toprule
        Margin ($\Delta$) & mAP & $\text{AP}_S$ & $\text{AP}_M$ & $\text{AP}_L$ \\
        \midrule
        0.00 & 33.6 & 16.2 & 35.3 & 39.6 \\
        % 0.01 & ---- & ---- & ---- & ---- \\
        \rowcolor{gray!20}
        0.05 & 34.9 & 15.0 & 36.4 & 40.5 \\
        0.10 & 33.2 & 13.2 & 34.6 & 40.2 \\
        % 0.20 & ---- & ---- & ---- & ---- \\
        \bottomrule
    \end{tabular}}\hspace{3mm}
% subfloat d
\subfloat[Impact of the local patch size on detection performance and latency. 
\label{tab:ablation_patch_size}]{
\tablestyle{10pt}{1.45}\begin{tabular}{cc |cccc c}
        \toprule
        Patch Size & $K$ & mAP & $\text{AP}_S$ & $\text{AP}_M$ & $\text{AP}_L$ & Latency (ms) \\
        \midrule
        $256 \times 256$ & 160 & 31.7 & 13.8 & 34.1 & 36.3 & 416 \\
        \rowcolor{gray!20}
        $512 \times 512$ & 40 & 34.9 & 15.0 & 36.4 & 40.5 & 357 \\
        $1024 \times 1024$& 10 & 30.1 & 14.7 & 31.2 & 33.4 & 353 \\
        \bottomrule
    \end{tabular}} 
% main caption
\label{tab:ablations}\vspace{-3mm}
\end{table*}

\textbf{Sensitivity of the Margin Hyperparameter ($\Delta$).} As detailed in Table~\ref{tab:ablations}(c), $\mathcal{L}_{LPM}$ achieves optimal performance at $\Delta = 0.05$. Setting $\Delta = 0.00$ ensures a basic local maximum but lacks a distinct numerical gap, yielding 33.6\% mAP. Introducing a positive margin of 0.05 effectively sharpens the Gain Map, providing precise spatial guidance for subsequent patch selection. Conversely, an excessively large margin ($\Delta = 0.10$, dropping to 33.2\% mAP) imposes overly strict penalties that disrupt the natural continuous distribution of spatial features. Consequently, we adopt $\Delta = 0.05$ as our default configuration.

\textbf{Impact of Local Patch Size.} As shown in Table~\ref{tab:ablations}(d), the $512 \times 512$ resolution emerges as the optimal sweet spot, delivering the best balance between detection accuracy and inference speed. Small patches ($256 \times 256$) fragment the scene and truncate objects, reducing mAP to 31.7\%, while the sheer volume of crops spikes latency to 416 ms. Conversely, large patches ($1024 \times 1024$) encompass vast uninformative backgrounds, diluting the sparse perception objective and degrading mAP to 30.1\%. Consequently, the $512 \times 512$ configuration provides the ideal spatial context, achieving peak accuracy while maintaining highly competitive efficiency.

\begin{table}[t]
\renewcommand{\arraystretch}{1.4}
    \centering
    \caption{Different patch selection strategies.}
    \label{tab:ablation_selection}
    \begin{tabular}{l cccc}
        \toprule
        Selection Strategy & mAP & $\text{AP}_S$ & $\text{AP}_M$ & $\text{AP}_L$ \\
        \midrule
        Rigid Suppression (NMS) & 30.0 & 12.3 & 30.6 & 35.8 \\
        Soft-Subtraction (Gaussian) & 33.4 & 14.1 & 35.2 & 40.1 \\
        \rowcolor{gray!20}
        Soft-Subtraction (Linear, Ours) & 34.9 & 15.0 & 36.4 & 40.5 \\
        \bottomrule
    \end{tabular}
\end{table}

\textbf{Effectiveness of Patch Selection Strategies.} As detailed in Table~\ref{tab:ablation_selection}, our linear Soft-Subtraction strategy significantly outperforms alternative mechanisms, achieving the optimal performance of 34.9\% mAP. The baseline Rigid Suppression (NMS) limits the overall accuracy to a suboptimal 30.0\% mAP. Transitioning to a Gaussian Soft-Subtraction kernel yields a substantial improvement (reaching 33.4\% mAP), demonstrating the clear advantage of applying spatial overlap penalties. Ultimately, our linear decay kernel proves superior. Unlike the centralized smoothing effect of a Gaussian kernel, the linear penalty accurately models the uniform geometric nature of physical patch overlaps. This precisely aligns with our set cover objective, maximizing object coverage while successfully avoiding redundant sampling.

\subsection{Visualization}

% \begin{figure}
%     \centering
%     \includegraphics[width=1.0\linewidth]{figures/coverage_vis.pdf}
%     % \caption{\textbf{Statistical analysis of object coverage rates.} The per-image distribution histogram (left) and cumulative distribution curve (right) demonstrate that our sparse selection strategy successfully captures the vast majority of target objects across the dataset.}
%     \caption{\textbf{Statistical analysis of object coverage rates.} (a) Per-image distribution histogram. (b) Cumulative distribution curve. Both charts demonstrate that our dynamic patch allocation effectively covers the vast majority of targets across the dataset.}
%     \label{fig:stats_coverage}
% \end{figure}

\begin{figure*}[!t]
    \centering
    \includegraphics[width=1.0\linewidth]{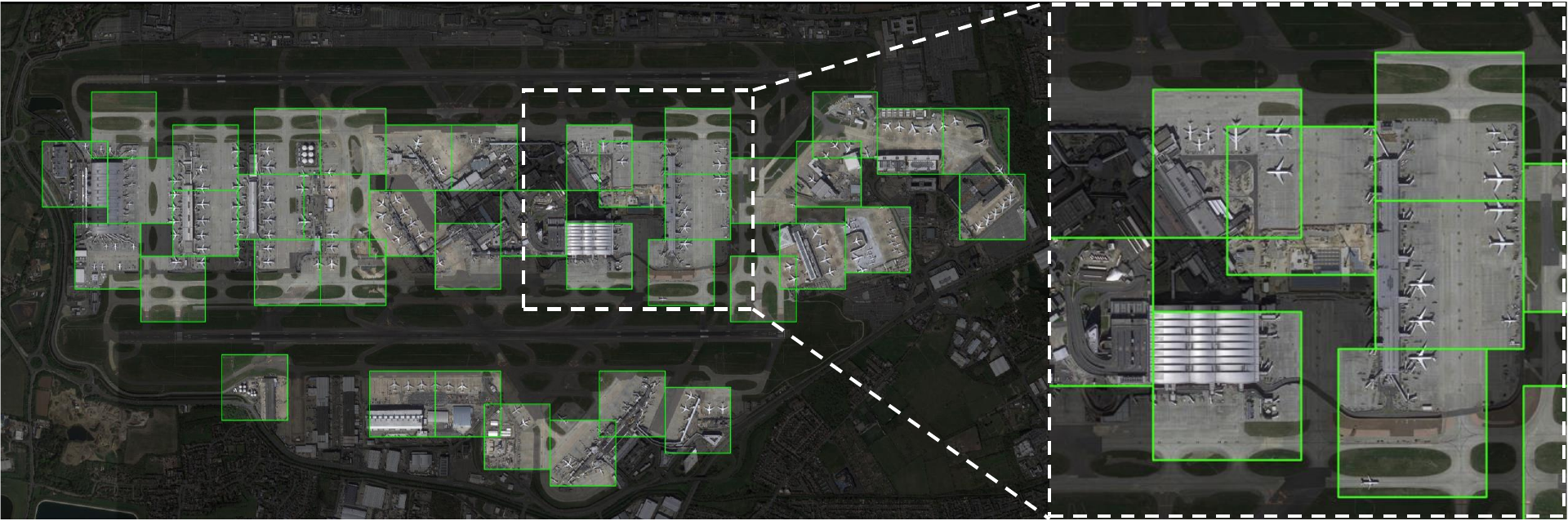}
    \caption{\textbf{Visualization of dynamic patch selection.} Green boxes indicate the selected local patches. Our method efficiently concentrates computational resources on object-dense regions while filtering out vast uninformative backgrounds. A zoomed-in view is shown on the right.}
    \label{fig:vis_patch1}
\end{figure*}

\begin{figure*}[t]
    \centering
    \includegraphics[width=1.0\linewidth]{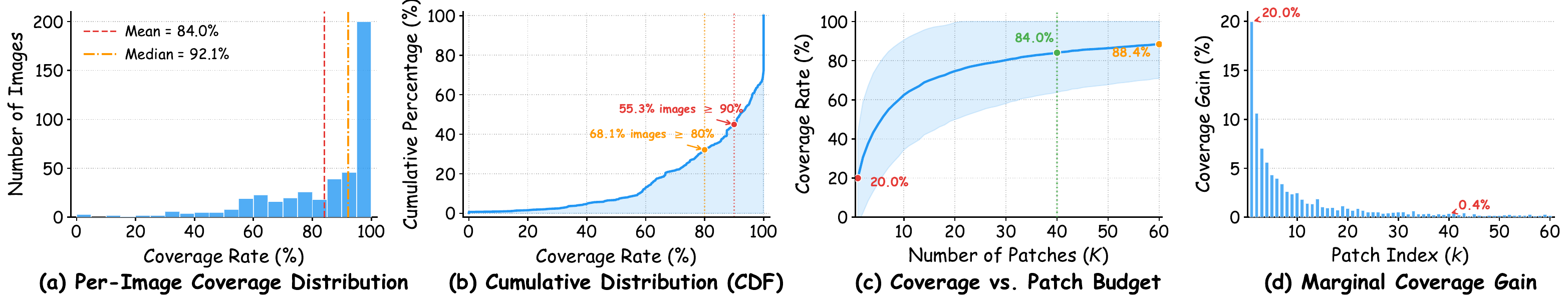}
    \caption{\textbf{Statistical analysis of object coverage rates.} \textbf{(a)} Per-image coverage distribution and \textbf{(b)} cumulative distribution curve (CDF) across the dataset. \textbf{(c)} Average coverage rate versus the patch budget ($K$). \textbf{(d)} Marginal coverage gain per patch iteration. These results collectively demonstrate that our dynamic allocation rapidly achieves high object coverage while efficiently managing spatial redundancy.}
    \label{fig:stats_coverage}
\end{figure*}

To validate the effectiveness of our sparse perception paradigm, we provide both qualitative and quantitative analyses. As visualized in Fig.~\ref{fig:vis_patch1}, our method dynamically concentrates high-resolution patches on dense object clusters while successfully bypassing vast uninformative backgrounds, intuitively demonstrating its capability to minimize spatial redundancy. Across the entire dataset, the statistical distributions in Fig.~\ref{fig:stats_coverage}(a) and (b) corroborate this robustness, confirming that our strategy consistently maintains high overall object coverage. More importantly, as illustrated in Fig.~\ref{fig:stats_coverage}(c) and (d), the coverage rate converges rapidly as the patch budget $K$ increases, accompanied by a sharp decay in marginal coverage gain. This perfectly aligns with the diminishing returns property of our geometric set cover formulation, proving that our soft-subtraction greedy algorithm efficiently maximizes object perception within a limited computational budget.

\begin{figure*}[!t]
    \centering
    \includegraphics[width=1.0\linewidth]{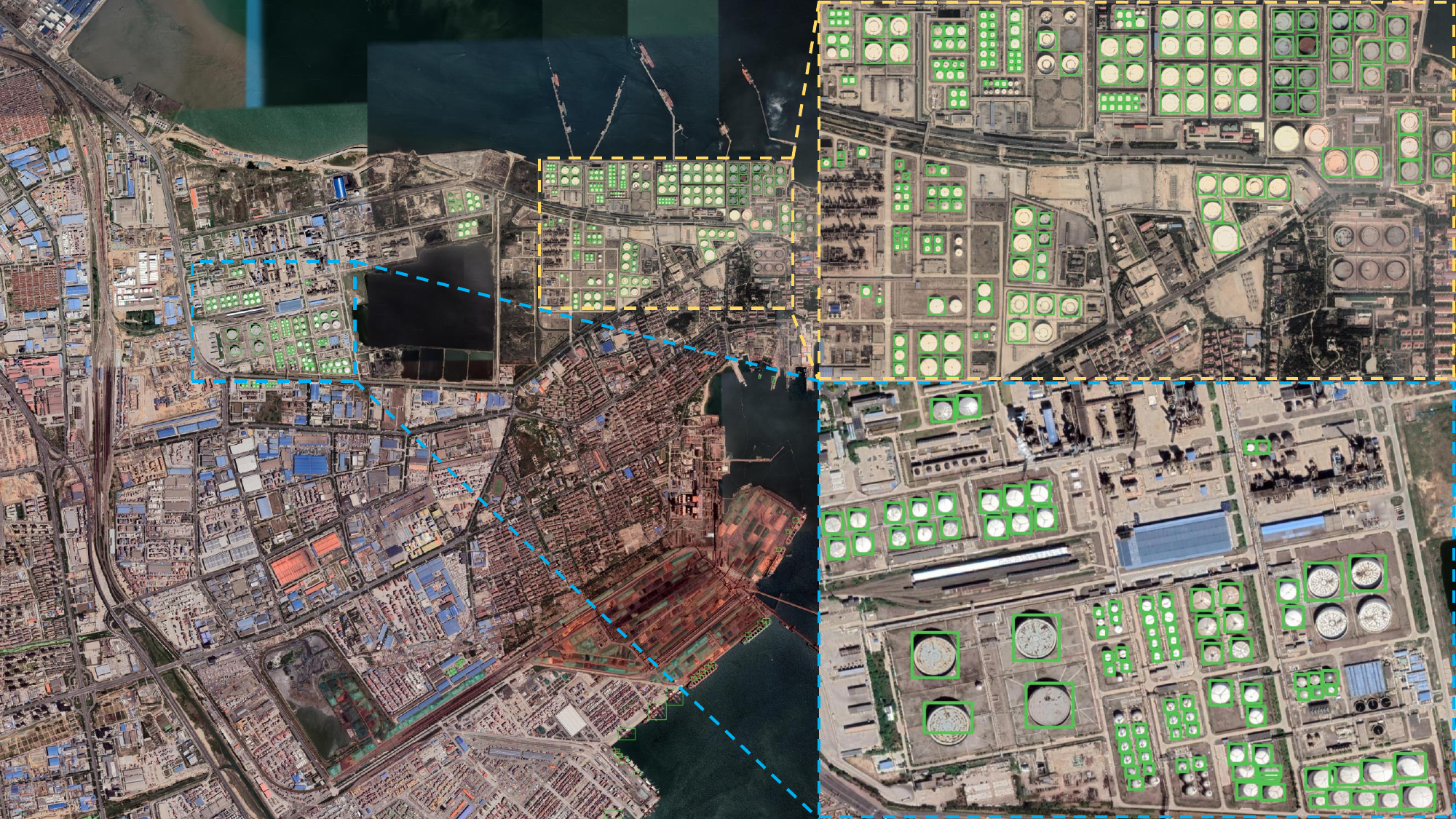}
    \caption{\textbf{Visualization of the final detection results in a massive UHR scenario.} The zoomed-in panels on the right demonstrate our method's capability to accurately localize densely packed small objects. By effectively leveraging macroscopic context, the framework maintains robust perception across the expansive background.}
    \label{fig:vis_dets}
\end{figure*}

Building upon this efficient patch extraction, Fig.~\ref{fig:vis_dets} presents the final end-to-end detection results. As highlighted by the zoomed-in panels, our method accurately localizes densely packed small objects within the expansive UHR imagery. By seamlessly integrating macroscopic scene awareness into the decoding process, our method ensures robust and precise perception across continuous wide-area scenarios.

\section{Discussions}
\label{discussion}
In this study, we propose UHR-DETR, an end-to-end transformer-based object detector designed to resolve the severe computational bottlenecks and the extreme contrast between vast backgrounds and small objects in massive ultra-high-resolution imagery. By formulating sparse feature extraction as a geometric set cover problem, our method dynamically selects informative patches using ISSGA guided by the LPM loss. Coupled with a Global-Local Decoupled Decoder, UHR-DETR successfully integrates macroscopic scene awareness with microscopic details, achieving an optimal trade-off between detection accuracy and inference speed.

To further clarify the design principles and inherent characteristics of our method, this section focuses on answering the following three key questions.

% \textit{-- Why use a Gain Map to model object distribution instead of traditional clustering algorithms or object-level density maps?}
\textit{-- Why use a Gain Map to model object distribution instead of coarse detection map or density maps?}

The fundamental advantage of the Gain Map lies in shifting the global perception objective from precise object localization to regional value estimation. Conventional coarse detection~\cite{pan_2022_cvpr, spdet_2023_icme} and density mapping~\cite{dmnet_2020_cvprw, saccadedet_2024_ecmlpkdd} paradigms assume that the network can achieve fine-grained localization (\textit{e.g.}, predicting preliminary bounding boxes or precise object centers) during the heavily downsampled global stage. However, in massive UHR scenarios, small object features are severely obliterated after heavy downsampling, rendering such early-stage localization highly unreliable. In contrast, our Gain Map bypasses this bottleneck. Operating on macroscopic, low-resolution features, it only evaluates the relative information richness of each sub-region (\textit{i.e.}, perceiving the expected quantity or value of objects within a local area) without demanding exact spatial coordinates. By aggregating coarse semantic responses to estimate regional value rather than individual object positions, this mechanism effectively tolerates weak initial signals, ensuring informative patches are reliably identified for subsequent high-resolution processing.

\textit{-- Why is macroscopic global context necessary when local patches already capture fine-grained object details?}

% In massive UHR scenes, small objects often exhibit extremely weak isolated feature responses. Relying solely on localized crops deprives the network of the broader scene context needed to resolve semantic ambiguities. The Global-Local Decoupled Decoder elegantly solves this by using Global Cross-Attention to inject macro-level structural cues into the localized decoding process. This ensures that while bounding box regression is precisely executed on high-resolution local features, the classification and target association are continuously informed by the global scene. Consequently, this holistic awareness significantly reduces false positives and strengthens the feature representation of isolated small objects that would otherwise be indistinguishable from background noise.

In massive UHR scenes, small objects often exhibit extremely weak isolated feature responses. Relying solely on localized crops deprives the network of the broader scene context needed to resolve semantic ambiguities. The Global-Local Decoupled Decoder solves this by using Global Cross-Attention to inject macroscopic structural cues into the localized decoding process. This mechanism fundamentally enhances the semantic representation and decisional certainty of isolated objects. As quantitatively demonstrated in Table~\ref{tab:ablations}(a), integrating this global context yields a 1.1\% improvement in overall mAP. More critically, by effectively resolving semantic ambiguities, it drives a substantial 2.1\% surge in small object precision ($\text{AP}_S$, from 12.9\% to 15.0\%).

\textit{-- What are the inherent limitations of the proposed UHR-DETR framework?}

% The primary limitation of UHR-DETR lies in its efficiency degradation when encountering homogeneously dense object distributions. While our dynamic patch allocation excels in typical UHR scenarios characterized by localized object clusters, its computational advantage diminishes in extreme failure cases where objects are uniformly scattered across the entire canvas. In such uniformly dense distributions, the absence of spatial sparsity forces the network to allocate a massive number of patches ($K_{infer}$) everywhere to satisfy the set cover objective. Consequently, the overhead of the dynamic routing mechanism, combined with the sheer volume of processed regions, causes the inference latency to increase significantly, eventually approaching that of exhaustive sliding-window baselines.

The primary limitation of UHR-DETR lies in its efficiency degradation when encountering homogeneously dense object distributions. While our dynamic patch allocation excels in typical UHR scenarios characterized by localized object clusters, its computational advantage diminishes in extreme failure cases where objects are uniformly scattered across the entire canvas. In such uniformly dense distributions, the absence of spatial sparsity forces the network to allocate a massive number of patches ($K_{\text{infer}}$) everywhere to satisfy the set cover objective. Consequently, the overhead of the dynamic routing mechanism, combined with the sheer volume of processed regions, causes the inference latency to increase significantly, eventually approaching that of exhaustive sliding-window baselines. To mitigate this, future work could explore adaptive resolution scaling to dynamically adjust processing resolutions based on regional difficulty, thereby maintaining efficiency even without spatial sparsity.

\section{Conclusion}
\label{conclusion}

% In this paper, we propose UHR-DETR, a highly efficient end-to-end framework tailored for ultra-high-resolution object detection. To overcome the computational bottlenecks of massive imagery, we introduce the Coverage-Maximizing Sparse Encoder, which formulates sparse feature extraction as a geometric set cover problem. Driven by the Iterative Soft-Subtraction Greedy Algorithm and the Local Peak Margin loss, this encoder dynamically allocates local patches to achieve maximum object coverage with minimal spatial redundancy. Furthermore, the Global-Local Decoupled Decoder effectively integrates macroscopic scene awareness into the localized decoding process within an end-to-end paradigm. Extensive experiments on the STAR and SODA-A benchmarks demonstrate that UHR-DETR achieves an optimal trade-off between detection accuracy and inference speed, delivering nearly a 10$\times$ speedup over traditional sliding-window baselines. While its computational advantage may diminish in extreme edge cases with uniformly dense object distributions, we believe this dynamic, globally-informed perception paradigm offers a scalable and practical foundation for real-world massive scene understanding.

In this paper, we propose UHR-DETR, an end-to-end framework designed to resolve the fundamental conflict between preserving fine-grained features for small objects and managing the prohibitive computational cost of ultra-high-resolution imagery. By dynamically extracting information-rich regions and synchronizing them with macroscopic context, our method achieves highly efficient sparse perception without relying on exhaustive sliding windows. Extensive experiments on the STAR and SODA-A datasets demonstrate that our approach strikes an optimal balance between detection accuracy and efficiency, yielding nearly a $10\times$ inference speedup over traditional baselines. However, the computational advantage of this sparse strategy naturally diminishes in extreme edge cases featuring uniformly dense object distributions across the entire canvas; future work could explore adaptive resolution scaling to mitigate such bottlenecks. We hope this work can provide a new paradigm and practical solutions for ultra-wide-area remote sensing object perception.

\section*{Acknowledgments}
This research was supported by the National Key Research and Development Program under Grant 2024YFF1401003 and the National Natural Science Foundation of China under Grant 62271355.

%{\appendices
%\section*{Proof of the First Zonklar Equation}
%Appendix one text goes here.
% You can choose not to have a title for an appendix if you want by leaving the argument blank
%\section*{Proof of the Second Zonklar Equation}
%Appendix two text goes here.}

 % argument is your BibTeX string definitions and bibliography database(s)
% \bibliography{IEEEabrv,../bib/paper}

\bibliographystyle{IEEEtran}
% \bibliography{IEEEabrv,bib/chang,bib/jinwang,bib/zrx,bib/all}
\bibliography{IEEEabrv,bib/all_cite}

\vfill

\end{document}